\documentclass[wcp]{jmlr}
\usepackage{subfiles}
\usepackage{mathtools}
\usepackage{algorithm}
\usepackage{algpseudocode}
\usepackage{longtable}% for long tables
\usepackage{siunitx}
\usepackage{symbols}
\usepackage{booktabs}
\usepackage{wrapfig}
\makeatletter
\let\Ginclude@graphics\@org@Ginclude@graphics 
\makeatother

\title[Sliced Wasserstein Variational Inference]{Sliced Wasserstein Variational Inference}

  \author{\Name{Mingxuan Yi} \Email{mingxuan.yi@bristol.ac.uk}\\
  \addr School of Mathematics, University of Bristol, UK
  \AND
  \Name{Song Liu} \Email{song.liu@bristol.ac.uk}\\
  \addr School of Mathematics, University of Bristol, UK
 }

\begin{document}

\maketitle

\begin{abstract}
Variational Inference approximates an unnormalized distribution via the minimization of \textit{Kullback-Leibler} (KL) divergence. Although this divergence is efficient for computation and has been widely used in applications, it suffers from some unreasonable properties. For example, it is not a proper metric, i.e., it is non-symmetric and does not preserve the triangle inequality. On the other hand, optimal transport distances recently have shown some advantages over KL divergence. With the help of these advantages, we propose a new variational inference method by minimizing sliced Wasserstein distance--a valid metric arising from optimal transport. This sliced Wasserstein distance can be approximated simply by running MCMC but without solving any optimization problem. Our approximation also does not require a tractable density function of variational distributions so that approximating families can be amortized by generators like neural networks. Furthermore, we provide an analysis of the theoretical properties of our method. Experiments on synthetic and real data are illustrated to show the performance of the proposed method.
\end{abstract}


\section{Introduction}

\begin{wrapfigure}[16]{l}{0.61\textwidth}
    \subfigure{\includegraphics[width=0.3\textwidth]{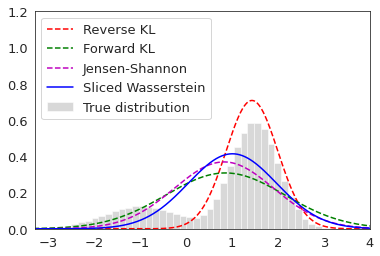}}
    \subfigure{\includegraphics[width=0.3\textwidth]{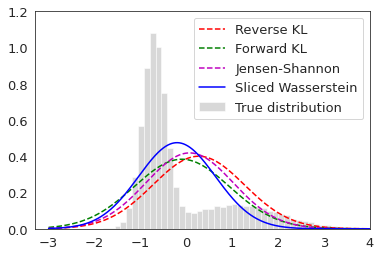}}
    \caption{The visualization of performances of using different discrepancies to match two probability distributions. \textit{Jensen-Shannon} divergence is a symmetric information divergence via averaging reverse and forward KL divergences.}
    \label{klandsw}    
\end{wrapfigure}

Variational inference (VI) is a method that recasts Bayesian inference as an optimization problem where it uses \textit{Kullback-Leibler} (KL) divergence as a measurement to capture the discrepancy of two probability distributions. Unlike the traditional inference methods that utilize Monte Carlo Markov Chains (MCMC) to sample from the target probability space,  VI is fast and lightweight in terms of computation. Therefore, it is preferred in many modern machine learning tasks.\\
\\
Optimal Transport (OT) \citep{villani2009optimal} has recently gained significant attentions in the machine learning community. Compared to KL divergence, OT gives a valid metric that is symmetric and preserves triangular inequality. It is reported to show good performances in some downstream applications \citep{arjovsky2017wasserstein, gulrajani2017improved}. While OT provides us with a new horizon on some old machine learning scenarios, the original OT problem requires a computationally demanding optimization procedure which impedes the popularity of applying the original methods. To address this difficulty, sliced Wasserstein distance \citep{bonnotte2013unidimensional, bonneel2015sliced} reduces the computational inefficiency of OT by projecting high dimensional probability distributions into univariate slices where OT problem has a closed-form solution. Similar to the standard Wasserstein distance,  sliced Wasserstein distance is still a valid metric function \citep{bonnotte2013unidimensional}. This metric has been successfully used in many practical tasks \citep{deshpande2018generative, kolouri2018sliced,kolouri2018sliced2} but it has not yet been studied in variational inference. See Figure \ref{klandsw} for the fitted Gaussian distributions (dotted and solid lines) obtained by minimizing sliced Wasserstein distance and various information divergences. It is known that VI tends to search modes with the reverse KL divergence but tends to spread the mass with the forward KL divergence. Sliced Wasserstein distance has a different behavior compared to three other information divergences. In Figure \ref{klandsw} (b), the approximating distribution is initialized to the right side and all information divergences are not as robust as sliced Wasserstein distance to fit the higher mode.

In this paper, we extend sliced Wasserstein distance to variational inference tasks. The distance between the variational and the target distribution is approximated using MCMC. Consequently, we sequentially minimize sliced Wasserstein distance between the variational distribution and the marginal distributions of MCMC. With MCMC evolves, such procedure guides the variational distribution to match the target under the minimization of a metric function. One advantage is that by leveraging the sliced Wasserstein distance, our method does not rely on simultaneous adversarial training \citep{mescheder2017adversarial,li2017approximate,zhang2020mcmc} to estimate the discrepancy but can still perform amortized inference \citep{gershman2014amortized}, i.e., use a parameterized function as a black-box sampler, e.g. neural networks, to capture target distributions. %Another similar work is contrastive divergence \cite{hinton2002training}, in which the gradient approximation is obtained with MCMC. However, our method uses MCMC to estimate the discrepancy between two distributions.

\textbf{Contributions:} i). We develop a new variational inference method that minimizes sliced Wasserstein distance -- a statistical distance. ii). The convergence and asymptotic properties of the proposed method are discussed as well as burn-in analysis. iii) We provide empirical studies to illustrate the method via experiments. 
\section{Background}
\subsection{Variational Inference}
Given an unnormalized probability distribution $\bar{\mu}(z)$, it could be difficult to obtain the normalizing constant $\beta=\int \bar{\mu}(z) dz$ and as such, we would not have assess to the true density $\mu(z) = \beta^{-1}\bar{\mu}(z)$. Variational inference aims to find a distribution $\upsilon_\phi(z)$ which approximates $\mu(z)$ as close as possible. Such approximations can be obtained via minimizing \textit{Kullback-Leibler} (KL) divergence  
\begin{equation}
    D_{KL} \big[\upsilon_\phi||\mu\big] = \int \upsilon_\phi(z) \log \frac{\upsilon_\phi(z)}{\mu(z)}dz \label{kl}
\end{equation} 
Note that $D_{KL} \big[\upsilon_\phi||\mu\big]=0$ if and only if $\upsilon_\phi(z) = \mu(z)$. However, Eq(\ref{kl}) is intractable because the density function $\mu(z)$ is known up to a normalizing constant. Instead, we can equivalently maximize \textit{Evidence Lower Bound} (ELBO)
\begin{equation}
    \log \beta \geq \mathcal{L}(\phi) = \mathbb{E}_{\upsilon_\phi(z)} \big[ {\log \bar{\mu}(z) - \log \upsilon_\phi (z)} \big] \label{kl-divergence}
\end{equation} 
Since we observe that the model evidence $\log \beta$ is a constant w.r.t. $\phi$ and the above inequality becomes tight if $D_{KL} \big(\upsilon_\phi||\mu\big)=0$. Optimization of ELBO requires the differentiation of the r.h.s. expectation. Gradient descent is a standard approach that allows for such optimization. To obtain a valid estimation of the gradient, a solution is to apply the score function method \citep{paisley2012variational, ranganath2014black}. An alternative solution to obtain the gradient of ELBO is the reparameterization trick \citep{kingma2013auto, rezende2014stochastic}. Vanilla VI leverages KL divergence but this can be substituted with any other $f$-divergences and importance sampling \citep{jerfel2021variational,wan2020f, prangle2019distilling} can be used to obtain gradient estimation for general $f$-divergences.

\subsection{Wasserstein Distance}
Wasserstein distance arises in optimal transport \citep{villani2009optimal} in which a distribution is transformed to another by moving probability mass. Wasserstein distance measures the cost of such a transformation. We denote $\mathcal{X}$ the sample space and let $\mathcal{Q}_p(\mathcal{X})$ be the set of Borel probability measures with finite $p$-th moment. Given two marginal distributions $\mu(x), \upsilon(y) \in \mathcal{Q}_p(\mathcal{X}) $, let $\Pi(\mu, \upsilon)$ be a set of any coupled joint distributions $\gamma(x, y)$ where $\int_\mathcal{X} \gamma(x, y)dx=\upsilon(y)$ and $\int_\mathcal{X} \gamma(x, y)dy=\mu(x)$. The $p-$Wasserstein distance is defined as 
\begin{equation}
\mathcal{W}_p(\mu, \upsilon) = \left\{\inf_{\gamma \in \Pi(\mu, \upsilon)} \int_{\mathcal{X} \times \mathcal{X}} \Vert{x-y}\Vert^p d \gamma(x, y)\right\}^{\frac{1}{p}} \label{wasserstein_primal}
\end{equation}
where $\Vert{x-y}\Vert$ is a cost function of moving a mass from $\mu$ to $\upsilon$. Intuitively, the $p-$Wasserstein distance aims to find an optimal joint distribution $\gamma(x, y)$ where the expected cost specified by Eq(\ref{wasserstein_primal}) achieves its minimum. Solving this optimization problem is generally difficult \citep{cuturi2013sinkhorn}, but we can rewrite $p-$Wasserstein distance in a univariate case as 
\begin{equation}
\begin{split}
    \mathcal{W}_p(\mu, \upsilon) &= \left\{\int_0^1 \Big|F^{-1}_{\mu}(t) - F^{-1}_{\upsilon}(t)\Big|^p dt\right\} ^{\frac{1}{p}} = \left\{\int_{\mathcal{X}} \Big|x - F^{-1}_{\upsilon}(F_\mu(x))\Big|^p d\mu(x)\right\} ^{\frac{1}{p}}
\end{split} \label{1d-wass}
\end{equation}
where $F(\cdot)$ is a cumulative distribution function and $F^{-1}(\cdot)$ is a quantile function of a probability distribution and the composition $F^{-1}_{v}F_u(\cdot)$ defines a transportation map that moves mass from $u(x)$ to $\upsilon(y)$. Given two empirical distributions, we can simply utilize Eq(\ref{1d-wass}) to estimate $p-$Wasserstein distance by sorting samples. 

\subsection{Sliced Wasserstein Distance}
Motivated by the computational efficiency of estimating Wasserstein distance with univariate distributions. We give a brief review of sliced Wasserstein distance \citep{bonneel2015sliced}. We first introduce \textit{Radon} transformation \citep{beylkin1984inversion}.\\
\\
Let $h(\cdot)$ be a function $h: \mathbb{R}^d \longrightarrow \mathbb{R}$. The \textit{Radon} transform is defined as 
\begin{equation}
    \mathcal{R}h_\theta(l) = \int_{S: l= \langle x, \theta\rangle} {h(x) d{S
    }} \label{sw-distance}
\end{equation}
 Eq(\ref{sw-distance}) defines a surface integral on a hyper-plane $S: l= \langle x, \theta\rangle $ where $l \in \mathbb{R}$ and $\theta \in \mathbb{S}^{d-1}$ where $\mathbb{S}^{d-1}$ is a unit ball embedded in $\mathbb{R}^d$. For any pair of vectors $\theta$ and $h$, we obtain a sliced function $h^R_\theta(\cdot)$. We note that marginalization of a high dimensional joint probability distribution can be regarded as a special case of the \textit{Radon} transform with $\theta=e_i$, where $e_i$ is an all-zero vector with only 1 at the $i$-th position. Note that the sliced function yielded by Eq(\ref{sw-distance}) is univariate. Leveraging this property, we define sliced Wasserstein distance for probability distributions $\mu(x)$ and $\upsilon(y)$ as the average distance resulting from these slices.
\begin{equation}
    \sw(\mu, \upsilon)=\left(\int_{\theta \in \mathbb{S}^{d-1}} \mathcal{W}_p^p(\mathcal{R}\mu_\theta, \mathcal{R}\upsilon_\theta) d\theta \right)^{\frac{1}{p}}
\end{equation}
Given an empirical distribution described by $\mu^n = \frac{1}{n} \sum_{i=1}^n \delta_{x_i}$, it is trivial to write down its \textit{Radon} transformation defined in Eq(\ref{sw-distance}) as $\mathcal{R}\mu^n_{\theta} = \frac{1}{n} \sum_{i=1}^n \delta_{\langle x_i, \theta\rangle}$.
We summarize the procedure of calculating sliced Wasserstein distance via empirical samples in \textbf{Algorithm 1}.

\begin{algorithm}
\floatconts
  {alg:sw_em}%
  {\caption{Estimation of Sliced Wasserstein Distance with Samples}}%

{%
\textbf{Require}: $\mu^n = \frac{1}{n} \sum_{i=1}^n \delta_{x_i}$ and $\upsilon^n = \frac{1}{n} \sum_{i=1}^n \delta_{y_i}$ (for simiplicity, we assume two distributions have the same number of observations)\\
\textbf{for} $k=0,1 \cdots m$
\begin{enumerate*}
  \item Sample $\theta_k$ from $\mathbb{S}^{d-1}$ uniformly,
  %\item Obtain slices $\mu^n^R_{\theta_k} = \frac{1}{n} \sum_{i=1}^n \delta_{\langle x_i, \theta_k\rangle}$ and $\upsilon^n^R_{\theta_k} = \frac{1}{n} \sum_{i=1}^n \delta_{\langle y_i, \theta_k\rangle}$
  \item Obtain slices and sort $\{\langle x_i, \theta_k\rangle\} \longrightarrow \{\langle x_j, \theta_k\rangle\}$ and $\{\langle y_i, \theta_k\rangle\} \longrightarrow \{\langle y_j, \theta_k\rangle\}$
  \end{enumerate*}
\Return $\sw(\mu^n,\upsilon^n) $ = $\left(\frac{1}{mn} \sum_{k=1}^{m} \sum_{j=1}^{n} \Big| \langle x_j, \theta_k\rangle - \langle y_j, \theta_k\rangle \Big| ^p\right)^{\frac{1}{p}}$
}%

\end{algorithm}
\section{Sliced Wasserstein Variational Inference}
We name our method as sliced Wasserstein variational inference (SWVI). We use the following notation: $\upsilon_\phi(z)$ as the variational distribution parameterized by $\phi$, and $\mu(z)= \beta^{-1} 
\bar{\mu}(z)$ as the target distribution, where $\beta$ is a normalizing constant. The problem we are interested in is finding an optimal parameter $\phi^*$ that minimizes sliced Wasserstein distance between the variational distribution and the target distribution. 
\begin{equation}
    \phi^* = {\arg\min}_{\phi}\ \sw(\mu, \upsilon_\phi ) \label{mde}
\end{equation}
Eq(\ref{mde}) defines a minimum distance estimator \citep{wolfowitz1957minimum} \citep{basu2011statistical} where we choose sliced Wasserstein distance as a specific metric function. In some applications, the density function of $\upsilon_\phi$ is not always tractable but we can still simulate samples from it and if $\mu$ also has a sampling distribution, we solve the following problem,
\begin{equation}
    \phi^* = {\arg\min}_{\phi}\ \sw(\mu^n, \upsilon^n_\phi )  \label{mde2}
\end{equation}
where $\mu^n$ and $\upsilon^n_\phi$ are sampling distributions of $\mu$ and $\upsilon_\phi$ with $n$ observations. The optimization problem defined in Eq(\ref{mde2}) does not require an explicit probability density function--allowing to design a more flexible variational distribution such as a neural network generator and a variational program \citep{ranganath2016operator}.

\subsection{Estimation of Sliced Wasserstein Distance}
Due to the intractability of the target distribution $\mu(z) = \beta^{-1} \bar{\mu}(z)$, the main idea of SWVI is to approximate sliced Wasserstein distance between the variational and the target distribution using MCMC and minimize it. Unlike variational inference, MCMC methods provide particle approximations by designing transition kernels of a Markov chain with invariant distribution $\mu(z)$.
Let $K(\cdot|\cdot)$ be a transition kernel of MCMC, and $\mu_0(z)$ be the initial distribution of the corresponding MCMC, e.g., a potential $\mu_0(z)$ can be chosen as $\upsilon_\phi(z)$. We denote by $\mu_t(z)$ the marginal distribution of the Markov chain after applying $t$ times transitions.
\begin{equation}
    \mu_t(z) = \int \mu_{t-1}(z')K(z|z')dz'%\text{,  where }\mu_0(z) = \upsilon_\phi(z)
\end{equation}
Given a sufficiently long run, $\mu_t(z)$ converges to $\mu(z)$ because of the stationary property of Markov chain. At the current stage, one can directly evaluate sliced Wasserstein distance $\sw(\mu, \upsilon_\phi)$ via 
\begin{equation}
    \sw(\mu, \upsilon_\phi) = \sw(\mu_t, \upsilon_\phi)\text{ as } t \to \infty
\end{equation}
Unfortunately, running a long enough MCMC chain is time consuming and it might be difficult to diagnose the burn-in period. To solve this problem, we instead evaluate a local distance $\sw(\mu_t, \upsilon_\phi)$ at every time step $t$ of iterating MCMC algorithms. Next we update parameters $\phi$ via gradient descent to minimize $\sw(\mu_t, \upsilon_\phi)$ per iteration. 
Since every $\mu_t(z)$ is an improvement of the previous $\mu_{t-1}(z)$, minimizing sliced Wasserstein distance guides the variational distribution $\upsilon_{\phi}(z)$ towards the target distribution $\mu(z)$. \\
\\Note that we use particle approximations to the marginal distribution $\mu_t$ by parallelizing $n$ Markov chains and we sample the same number of particles from the variational distribution $\upsilon_\phi$. Hence, the optimization problem is replaced with minimizing the sampled-based sliced Wasserstein distance $\sw(\mu^n_t, \upsilon^n_\phi)$. Therefore, SWVI defines the following sequential optimization problem 
\begin{equation}
    \{\min_{\phi} \sw(\mu^n_t, \upsilon^n_\phi)\}_{t \in \mathbb{N}}
\end{equation}
Heuristically, we also inherit the parameter from the previous iteration such that $\phi_t = \arg\min_{\phi} \sw(\mu^n_t, \upsilon^n_{\phi_{t-1}})$. We summarize this procedure in \textbf{Algorithm 2}. 

\begin{algorithm}
\floatconts
  {alg:swvi}%
  {\caption{Sliced Wasserstein Variational Inference (SWVI)}}%
  
{%
\textbf{Require:} An unnormalized probability distribution $\bar{\mu}(z)$, a Markovian transition kernel $K(z|z')$ with invariant distribution $\mu(z)$ and initial distribution $\mu_0(z)$, variational distribution $\upsilon_{\phi}(z)$. \\
\textbf{Initialize} $\phi_0 = \phi$ and $\mu^n_0(z)$ by sampling $n$ particles from $\mu_0(z)$\\
\textbf{for } $t=1, 2 \cdots T$
\begin{enumerate*}
  \item  Apply transition kernel $K(z|z')$ to $\mu^n_{t-1}$ to get $\mu^n_{t}$
  \item $\phi_{t} = \arg\min_{\phi} \sw(\mu^n_{t}, \upsilon^n_{\phi_{t-1}})$
  \end{enumerate*}
  \textbf{return } $\upsilon_{\phi_T}(z)$
}%
\end{algorithm}

%This procedure yields a sequence $\{\phi_m\}_{m=0,1,2\cdots}$(to be continued)

\subsection{Existence, Convergence and Consistency of SWVI}
It is important to understand the convergence of SWVI while the algorithm iterating with time $t$ and the asymptotic property if the number of Markov chains $n$ goes to infinity. Specifically, we show that SWVI converges if the Markov chain converges to the invariant distribution (\textbf{Theorem 1}) and SWVI is consistent (\textbf{Theorem 2}) if the number of Markov chains goes to infinity under some mild regularities. Inspired by \citep{bernton2019parameter} \citep{nadjahi2019asymptotic}, we make the following assuamptions\\\\
\textbf{Assumption A.1.} For any sequence of probability measures $\{\mu_t\}_{t\in \mathbb{N}}$, e.g., marginal probability densities yield by Markov chains, $\{\mu_t\}_{t\in \mathbb{N}}$ converges in sliced Wasserstein distance to $\mu$, i.e. $\lim_{t \to \infty} \sw(\mu_t, \mu ) = 0 $, $\mathbb{P}$-almost surely.\\
\textbf{Assumption A.2.} The map $\phi \to \upsilon_\phi$  is continuous, i.e., $\lim_{k\to \infty} ||\phi_k - \phi||=0$ implies weak convergence of $\upsilon_{\phi_k}$ to $\upsilon_\phi$.\\
\textbf{Assumption A.3.} For any data generating processes $\{\mu_{t}^{n}\}_{n\in \mathbb{N}}$, we have $\lim_{n \to \infty} \sw(\mu_{t}^{n}, \mu_t ) = 0 $. For $\lim_{n\to \infty} ||\phi_n - \phi||=0$, we have $\lim_{n \to \infty} \sw(\upsilon_{\phi_n}^{n}, \upsilon_\phi ) = 0 $, $\mathbb{P}$-almost surely.\\
\textbf{Assumption A.4.} For some $\epsilon > 0$ with $\epsilon^* = \inf_{\phi } \sw(\mu_t, \upsilon_\phi )$, $B_{\epsilon} = \{\phi \in \mathcal{H}: \sw(\mu_t, \upsilon_\phi) \leq \epsilon^* + \epsilon \}$ is bounded.\\\\
Note that all infimum are taken with $\phi \in \mathcal{H}$ where $\mathcal{H}$ is a parameteric space.  \textbf{A.1} indicates the convergence of Markov chain in the metric space. Indeed, it also implies weak convergence $\{\mu_t\}_{t\in \mathbb{N}} \to \mu$ in $\mathcal{Q}_p(\mathcal{X})$. This is a corollary from \citep{nadjahi2020statistical} where if $p$-Wasserstein distance metrizes weak convergence then the sliced $p$-Wasserstein metrizes weak convergence as well. We refer to \citep{villani2009optimal} for the study of weak convergence with Wasserstein distance. A straightforward result is that Langevin dynamic is a gradient flow of KL divergence in Wasserstein Space \citep{jordan1998variational} and \citep{liu2019understanding} generalizes it to other MCMC algorithms. 

\begin{theorem}
Under \textbf{Assumption A.1} and \textbf{A.2},
\begin{equation}
    \lim_{t \to \infty} \inf_{\phi} \sw(\mu_t, \upsilon_\phi )  = \inf_{\phi} \sw(\mu, \upsilon_\phi ) 
\end{equation}
\end{theorem}

\begin{theorem}
Under \textbf{Assumption A.2}, \textbf{A.3} and  \textbf{A.4}
\begin{equation}
    \lim_{n \to \infty} \inf_{\phi} \sw(\mu^n_t, \upsilon^n_\phi )  = \inf_{\phi} \sw(\mu_t, \upsilon_\phi ) 
\end{equation}
and the existence of minimum
$$
     \arg\min_{\phi}  \sw(\mu_t, \upsilon_\phi ) \neq \emptyset
$$
\end{theorem}

\begin{corollary}
If Markov chains evolve with time and the number of chains goes to infinity, we have
\begin{equation}
    %\lim_{t \to \infty}\lim_{n \to \infty} \inf_{\phi} \mathbb{E}\left[\sw(\mu^n_t, \upsilon^n_\phi ) \right] = \inf_{\phi} \sw(\mu, \upsilon_\phi ) 
    \lim_{t \to \infty}\lim_{n \to \infty} \inf_{\phi} \sw(\mu^n_t, \upsilon^n_\phi ) = \inf_{\phi} \sw(\mu, \upsilon_\phi )
\end{equation} 
\end{corollary} 
\textbf{Theorem 1} indicates that the sequential optimization problem of SWVI converges. The proof is analogous to \citep{nadjahi2019asymptotic} where the sequence of probability measures here is constructed by Markov chains. \textbf{Theorem 2} is similar to Corollary 6.11  by \cite{villani2009optimal} where if $\mu^n$ and $\upsilon^n$ weakly converge in $\mathcal{Q}_p(\mathcal{X})$ to $\mu$ and $\upsilon$, then $\lim_{n \to \infty} \mathcal{W}_p(\mu^n, \upsilon^n) = \mathcal{W}_p(\mu, \upsilon)$. \textbf{Corollary 3} is a direct result from \textbf{Theorem 1} and \textbf{2}.
Here we prove the infimum of the sequence $\{\sw(\mu^n_t, \upsilon^n_\phi )\}_{n\in\mathbb{N}}$ converges to the infimum of $\sw(\mu_t, \upsilon_\phi )$ in Appendix A.

\subsection{Burn-in Diagnosis}
SWVI utilizes MCMC to explore the target space. Hence the convergence of the corresponding Markov chain influences the accuracy of the variational approximation. Burn-in diagnosis of MCMC is a hard problem in the machine learning and statistics community. Several attempts are proposed to assess the convergence such as estimating marginal densities and using couplings to estimate an upper bound of Wasserstein distance \citep{biswas2019estimating}.
\\
\\
We show that our method does not suffer from the problem of determining the burn-in period of Markov Chain. Sliced Wasserstein distance itself can monitor the convergence of MCMC. Without of the loss of generality, we assume that the burn-in period $t<M$ for some $M\in \mathbb{N}$ such that,
\begin{equation}
\sw(\mu_{t+1}, \upsilon_\phi) \approx \sw(\mu_{t}, \upsilon_\phi)\approx \sw(\mu, \upsilon_\phi)\text{, for } t \geq M, \text{for all } \phi
\end{equation}
This implies that  $\min_{\phi} \sw(\mu_{t}, \upsilon_\phi)$ becomes stationary for $t \geq M$.
That is to say once the loss function achieves the minimum with a tolerance, SWVI congerves. We show the loss curves of our method of fitting 2D Gaussian distributions (corresponds to experiment \textbf{4.1}) in Figure 2 and compare it with 2-Wasserstein distance where it is tractable (we use sinkhorn divergence \cite{cuturi2013sinkhorn} as approxiamted 2-Wasserstein distance for mixture distributions).

\begin{figure}[h]
    \subfigure{\includegraphics[width=0.2675\textwidth]{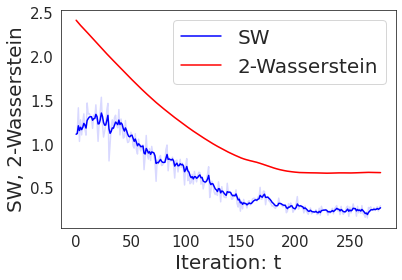}}
    \subfigure{\includegraphics[width=0.233\textwidth]{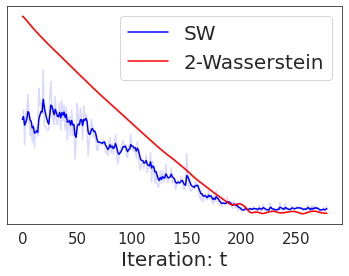}}
    \subfigure{\includegraphics[width=0.233\textwidth]{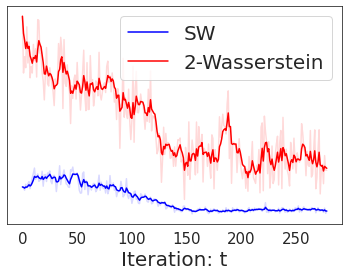}}
    \subfigure{\includegraphics[width=0.233\textwidth]{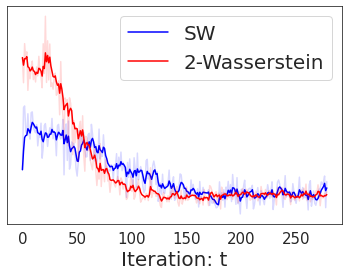}}
    \caption{Loss curves of fitting Gaussian distributions. (a) Mean-field Gaussian approximations.  (b)-(d) Full Gaussian approximations where (c) and (d) has a target bi-modal Gaussian mixtures. Compared with 2-Wasserstein (Sinkhorn approximations in (c) and (d)) distances between Gaussians. SWVI curves have empirically similar convergence rates to the convergence described by 2-Wasserstein distances}
\end{figure} \label{loss-cur}

%We refer this to a future work of burn-in diagnosis 

\subsection{Stochastic Optimization}
The sampled-based sliced Wasserstein distance $\sw(\mu^n_t, \upsilon^n_\phi )$ can be estimated by drawing samples respectively from $\mu_t(z)$ (via MCMC) and $\upsilon_{\phi}(z)$. Suppose that $\{z_i\}_{i=1,2\cdots n} \sim \upsilon_{\phi}(z)$ and $\{z'_i\}_{i=1,2\cdots n} \sim \mu_t(z)$. Sliced Wasserstein distance is then approximated by 
\begin{equation}
    \sw(\mu^n_t, \upsilon^n_{\phi}) \approx \mathcal{L}(\{z_i\}, \{z'_i\}) \label{app_sw}
\end{equation}
Here we rewrite the approximate distance as a function $\mathcal{L}(\cdot, \cdot)$ of two sets of samples according to \textbf{Algorithm 1}. In order to optimize the parameter of the variational distribution $\upsilon_\phi(z)$, we still need to reparameterize samples $\{z_i\}_{i=1,2\cdots n}$. This can be done by an amortized sampler that can be either a parametric probability distribution or a flexible neural network generator. The amortized sampler is written as $z(\phi) = g_\phi(\epsilon), \text{  } \epsilon \sim r(\epsilon)$, where $r(\epsilon)$ is a noise distribution and $g_\phi$ is a parametric model. We can use chain rule to obtain the gradient estimation of Eq(\ref{app_sw}),
\begin{equation}
    \nabla_\phi \mathcal{L}(\{z_i\}, \{z'_i\}) = \sum_{i=1}^{n} \nabla_{z_i}\mathcal{L}(\{z_i\}, \{z'_i\})\nabla_\phi z_i(\phi).
\end{equation}
This can be implemented easily via back-propagation. We also find it is helpful to warm-up MCMC by a few steps and then applies SWVI iteration. The only difference of such lagged procedure is that SWVI starts at a better initial distribution of MCMC. The practical implementation of SWVI is summarized in \textbf{Algorithm 3}
\begin{algorithm}
\floatconts
  {alg:swvi2}%
  {\caption{Practical Implementation of SWVI}}%
  
{%
\textbf{Require:} An unnormalized probability distribution $\bar{\mu}(z)$, a Markovian transition kernel $K(z|z')$ with invariant distribution $\mu(z)$ and initial distribution $\mu_0(z)$, variational distribution $\upsilon_{\phi}(z)$, learning rate $\alpha$ and warm-up lag $L$.  \\
\textbf{Initialize}  $\phi_0 = \phi$ and $\mu^n_0(z)$ by sampling $n$ particles from $\mu_0(z)$\\
\textbf{for} $t=1, 2, \cdots L-1$
\begin{enumerate*}
    \item Only run MCMC, i.e., apply kernel $K(z|z')$ to $\mu^n_{t-1}$ to get $\mu^n_{t}$
\end{enumerate*}
\textbf{for} $t=L, L+1 \cdots T$
\begin{enumerate*}
  \item Apply transition kernel $K(z|z')$ to $\mu^n_{t-1}$ to get $\mu^n_{t}$ with corresponding particles $\{z'_i\}_{i=1,2\cdots n} \sim \mu_t(z)$
  \item Draw $\{z_i\}_{i=1,2\cdots n} \sim \upsilon_{\phi_t}(z)$ with reparameterization
  \item Update parameter $\phi_{t+1} = \phi_t - \alpha \nabla_\phi \mathcal{L}(\{z_i\}, \{z'_i\})$
  \end{enumerate*}
  \textbf{return } $\upsilon_{\phi_T}(z)$
}%
\end{algorithm}

\subsection{Related Work}
To the best of our knowledge, we first introduce sliced Wasserstein distance into variational inference.
Sliced Wasserstein has been widely studied in generative modeling \citep{deshpande2018generative, kolouri2018sliced2, bonet2021sliced} as a measurement of the discrepancy of model distributions and data. \cite{ambrogioni2018wasserstein} combines orginal Wasserstein distances with variational inference where they introduce a new class of discrepancies that includes $f$-divergence allowing for variational inference problems. On the other hand, many alternative objectives to (reverse) KL divergence has been proposed in VI problems. For example, Stein discrepancy \citep{ranganath2016operator, liu2016stein}, forward KL divergence \citep{jerfel2021variational, prangle2019distilling, bornschein2014reweighted, dieng2017variational}, $\alpha$- divergence \citep{li2016renyi} and \textit{f}-divergences \citep{wan2020f}. Since our methods uses MCMC to estimate the metric function. This is similar to some previous works utilize the advantages of MCMC methods to improve variational inference. The authors of \citep{ruiz2019contrastive} proposes to use MCMC samples to estimate a new objective function to ELBO. The authors of \citep{naesseth2020markovian} estimates the gradient of forward KL divergence via runing MCMC. Similar to our work, The authors of \citep{li2017approximate} uses 'teacher-student' framework where MCMC samples teach the variational distribution to how to improve via minimizing different objectives. Another line of research to enrich the approximating families of VI is normalizing flows \citep{rezende2015variational}. Compared to our method, normalizing flows require manually design a bijective function whereas SWVI can use a simple non invertible neural net.

\section{Experiments}
For all experiments, we use sliced 1-Wasserstein distance. Details of experiment settings can be found in Appendix B.
\subsection{Toy Experiment}

In this experiment, we set target distributions as a 2D Gaussian distribution and a bi-modal Gaussian mixture.  We fit the variational distribution to the target distribution via vanilla variational inference under reverse KL divergence and the proposed method SWVI. In our method, we adopt the random walk Metropolis-Hastings algorithm as our MCMC instance. 
\begin{figure}[h]
\includegraphics[width=0.24\textwidth]{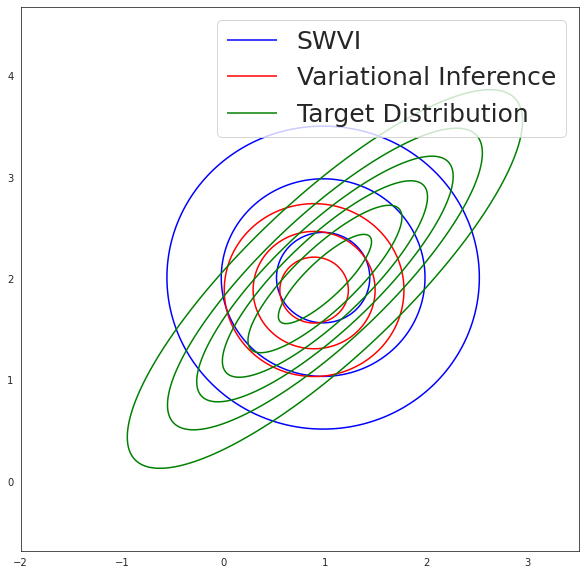}
\hspace{0ex}
\includegraphics[width=0.24\textwidth]{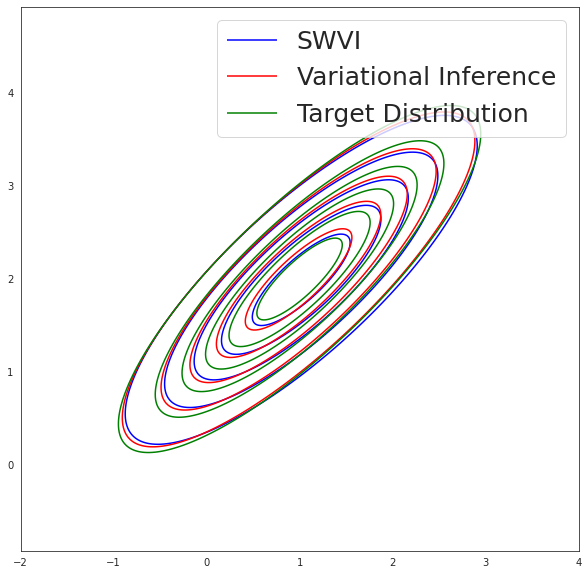}
\hspace{0ex}
\includegraphics[width=0.24\textwidth]{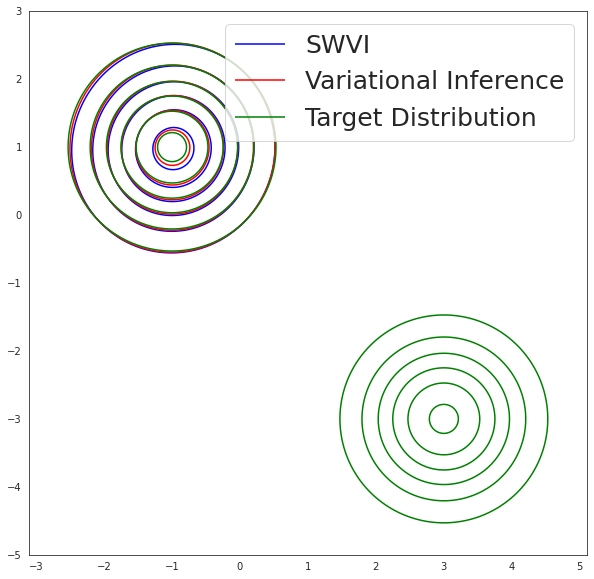}
\hspace{0ex}
\includegraphics[width=0.24\textwidth]{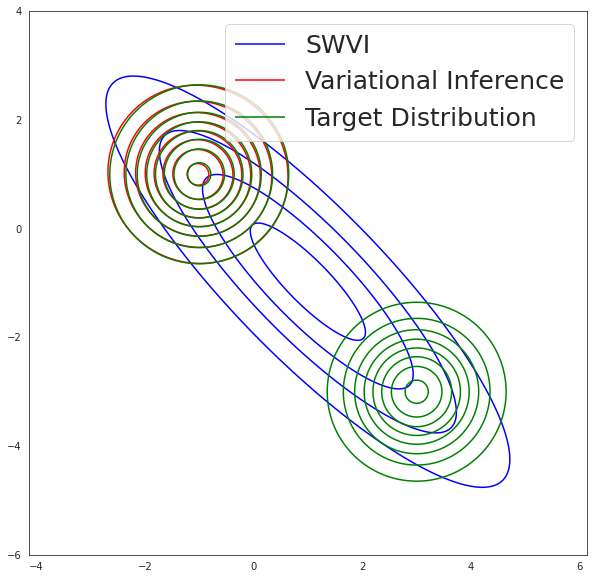}
\caption{Approximating 2D synthetic distributions with VI and SWVI}
\vspace*{-5mm}
\label{fig:2d-exp}
\end{figure}\\\\
In 1st figure of Figure \ref{fig:2d-exp}, we use a mean-field Gaussian distribution as the variational distribution and it shows that SWVI results in an approximation with a larger variance compared to standard VI. In the 2nd figure, we use a regular Gaussian distribution with fully trainable covariance matrix, both VI and SWVI can approximate the target distribution well. For the 3rd and 4th figures, the target distribution is set as a Gaussian mixture model, VI always fits one mode but SWVI can have different behaviours if we choose different step sizes in MCMC (random walk with standard deviations 0.2 and 2.5).

\subsection{Implicit Variational Distribution with Neural Nets}
\begin{wrapfigure}[10]{h}{0.5\textwidth}
\vspace*{-8mm}
  \begin{center}
\includegraphics[width=.24\textwidth]{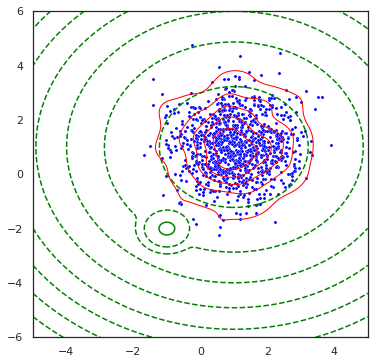}
\includegraphics[width=.24\textwidth]{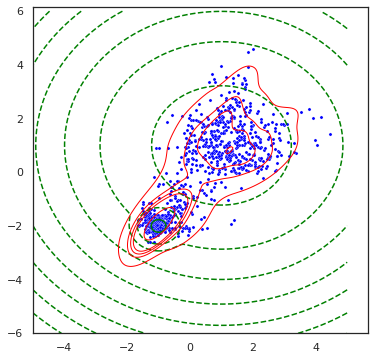}
  \end{center}
  \vspace*{-10mm}
  \caption{Fitting a mixture distribution with SVGD (L) and SWVI (R)}
\vspace*{-5mm}
\end{wrapfigure}
We show an experiment where we have amortized SWVI such that the variational distribution is a neural network generator without an explicit density function. Note that the proposed method in \textbf{Algorithm 3} does not require a closed form of density function of $\upsilon_\phi(z)$. Hence, we can easily adapt a neural net that can generate samples from a more flexible distribution. For comparison, we also implemented amortized Stein variational gradient (SVGD) method \citep{liu2016stein,feng2017learning}. 
The target distribution is a mixture of two Gaussians but one with a larger variance and another with a smaller variance. We fit a neural net generator to this distribution with SWVI and amortized SVGD. This experiment shows that amortized SVGD fails to capture the other mode. The reason is that the kernel function (RBF) cannot adjust the bandwidth to the two modes with different ranges of variance. However, with the asymptotic guarantees of MCMC, the generator trained with SWVI can efficiently capture two different modes and outputs considerably better samples.

\subsection{Bayesian Logistic Regression}
We apply SWVI to binary classification tasks in the UCI repository \citep{asuncion2007uci} with Bayesian logistic regression models. We set the prior distribution to be a constant for the Bayesian logistic model and Langevin Dynamics (with learning rate 0.0001) \citep{brooks2011handbook, welling2011bayesian} as the MCMC instance. The variational distributions are both mean-field Gaussians for SWVI and VI. We present results in Table 1 and it can be seen that the performance of SWVI is on par with the vanilla VI. 
 \begin{table}[h]
\begin{center}
\begin{tabular}{ c|c|c }
\hline
Dataset & Mean-field SWVI & Mean-field VI \\
\hline
Heart & 0.852$\pm$0.019 & \textbf{0.855}$\pm$0.030 \\
Wine & 0.716$\pm$0.025 & \textbf{0.731}$\pm$0.012 \\
Ionosphere &\textbf{0.771}$\pm$0.071 &  0.767$\pm$0.062  \\
\hline
\end{tabular}
\caption{\label{tab:table-name}Test accuracy for Bayesian logistic regression (32 posterior samples are used).}
\end{center}
\label{table:table1}
\end{table}
\subsection{Latent Generative Modeling}
We consider a latent generative model where observations $x$ is supposed to be generated by a latent variable $z$. In this model, $\theta$ parameterizes  a  likelihood function $p_\theta(x|z)$. Given a prior $p(z)$ for the latent variable, we have a posterior distribution
\begin{equation}
    \mu_\theta(z|x) \propto p(z) p_\theta(x|z)
\end{equation}
For a given likelihood function, VI can be used to obtain an approximated posterior distribution $\upsilon_\phi(z)$ via minimizing $D_{KL} \big[\upsilon_\phi||\mu_\theta\big]$. Suggested by \cite{hoffman2017learning}, there exists a approximation gap with VI such that MCMC can refine the latent variable $z$ sampled from the approximated posterior $\upsilon_\phi(z)$ to get a higher completed data log-likelihood. We compare the posterior approximations with VI, VI+MCMC \citep{hoffman2017learning} and SWVI on MNIST data set and report the completed data log-likelihood as evaluation of the approximations. For the likelihood function $p_\theta(x|z)$, we use a decoder from a variational auto-encoder pre-trained on training set. We evaluate the posterior approximations on 100 samples from the test set. For the variational distributions, we use mean-field Gaussian for both VI and SWVI and we also use an implicit variational distribution with neural nets for SWVI. The MCMC method used is Hamiltionian Monte Carlo (HMC) \citep{neal2011mcmc} More details about the experiment settings can be found in Appendix.

%\begin{equation}
    %\begin{split}
     %   p(z) &\sim N (0, I) \\
    %    p_\theta(x|z) &\sim Bernoulli\left[\sigma(Wz+b)\right]
   % \end{split}
%\end{equation}
%Such that we obtain an approximated posterior $\upsilon_\phi(z)$ and a likelihood $p_\theta(x|z)$ by maximizing the ELBO
%\begin{equation}
    %\log p_\theta(x) \geq \mathbb{E}_{\upsilon_\phi(z)} \big[ {\log p(z) p_\theta(x|z) - \log \upsilon_\phi (z)} %\big]
%\end{equation}
% where 
%\begin{equation}
   %\log p_\theta(x) = \log \int \frac{p(z) p_\theta(x|z)}{\upsilon_\phi (z)}\upsilon_\phi (z) dz
%\end{equation}
%We next fix the likelihood function $p_\theta(x|z)$ and apply HMC to refine samples from the posterior $q_\theta(z|x)$.
% We apply SWVI to this latent variable model and compare it with standard VI and 'VI+HMC' refinement.
\begin{table}[h]
\begin{center}
\begin{tabular}{ c|c }
\hline \\
Method & Complete Data Log-Likelihood \\
\hline
Mean-field VI &   -113.78$\pm$0.18\\
Mean-field VI+HMC &  -113.53$\pm$0.27 \\
Mean-field SWVI &  -113.62$\pm$0.20\\
Neural net SWVI &  \textbf{-113.16}$\pm$0.31\\
\hline
\end{tabular}
\caption{\label{tab:table-name}Comparisons of posterior approximations (10 posterior samples are used to calculate standard deviations). We observe that SWVI with implicit neural net distribution achieves the high log-likelihood.}
\end{center}
\label{table:table2}
\end{table}
%The advantage of using SWVI is that we can obtain a refined posterior distribution than that obtained with standard VI and no MCMC is needed further for sampling latent variables that allows for end-to-end modelling.
%\begin{figure}
%\includegraphics[width=0.49\textwidth]{a1.png}
%\includegraphics[width=0.49\textwidth]{a2.png}
%\includegraphics[width=0.49\textwidth]{a3.png}
%\caption{Generated Pictures}
%\vspace*{-5mm}
%\label{mnist}
%\end{figure}

\section{Conclusion}
We introduced sliced Wasserstein variational inference--a new method of variational inference to minimize a sequence of discrepancies between the variational distribution and the target distribution. SWVI utilizes MCMC to construct such sequence where at each iteration the variational distribution is improved towards the target. We also provide an analysis of the theoretic guarantees to the convergence of the proposed method and justify the consistency when the number of Markov chains goes to infinity. SWVI is flexible where the approximation can be either an implicit black-box sampler or a standard parametric probability distribution. We illustrate the performance of SWVI on several experiments. SWVI is a general algorithm that applies to density fitting and Bayesian learning problems. In latent generative modeling, SWVI can be used to refine the posterior distributions such that the log likelihood can be improved. In future work, we will study the asymptotic distribution of the estimator obtained via SWVI and its the finite-sample behavior as well. In addition, more practical applications will be further studied such as deep generative modelings and image inpainting. 
%A figure in Fig.~\ref{fig:spiral}. Please use high quality graphics for your camera-ready submission -- if you can use a vector graphics format such as \texttt{.eps} or \texttt{.pdf}.

%An example of citation~\cite{DBLP:conf/acml/2009}.

%\acks{Acknowledgements should go at the end, before appendices and references. You can uncomment this for the camera-ready version on paper acceptance.}

%\bibliographystyle{plain}
\bibliography{acml22}

\begin{thebibliography}{45}
\providecommand{\natexlab}[1]{#1}
\providecommand{\url}[1]{\texttt{#1}}
\expandafter\ifx\csname urlstyle\endcsname\relax
  \providecommand{\doi}[1]{doi: #1}\else
  \providecommand{\doi}{doi: \begingroup \urlstyle{rm}\Url}\fi

\bibitem[Ambrogioni et~al.(2018)Ambrogioni, G{\"u}{\c{c}}l{\"u},
  G{\"u}{\c{c}}l{\"u}t{\"u}rk, Hinne, van Gerven, and
  Maris]{ambrogioni2018wasserstein}
Luca Ambrogioni, Umut G{\"u}{\c{c}}l{\"u}, Ya{\u{g}}mur
  G{\"u}{\c{c}}l{\"u}t{\"u}rk, Max Hinne, Marcel~A van Gerven, and Eric Maris.
\newblock Wasserstein variational inference.
\newblock \emph{NeurIPS}, 2018.

\bibitem[Arjovsky et~al.(2017)Arjovsky, Chintala, and
  Bottou]{arjovsky2017wasserstein}
Martin Arjovsky, Soumith Chintala, and L{\'e}on Bottou.
\newblock Wasserstein generative adversarial networks.
\newblock In \emph{ICML}, 2017.

\bibitem[Asuncion and Newman(2007)]{asuncion2007uci}
Arthur Asuncion and David Newman.
\newblock Uci machine learning repository, 2007.

\bibitem[Basu et~al.(2011)Basu, Shioya, and Park]{basu2011statistical}
Ayanendranath Basu, Hiroyuki Shioya, and Chanseok Park.
\newblock \emph{Statistical inference: the minimum distance approach}.
\newblock CRC press, 2011.

\bibitem[Bernton et~al.(2019)Bernton, Jacob, Gerber, and
  Robert]{bernton2019parameter}
Espen Bernton, Pierre~E Jacob, Mathieu Gerber, and Christian~P Robert.
\newblock On parameter estimation with the wasserstein distance.
\newblock \emph{Information and Inference: A Journal of the IMA}, 8\penalty0
  (4):\penalty0 657--676, 2019.

\bibitem[Beylkin(1984)]{beylkin1984inversion}
Gregory Beylkin.
\newblock The inversion problem and applications of the generalized radon
  transform.
\newblock \emph{Communications on pure and applied mathematics}, 37\penalty0
  (5):\penalty0 579--599, 1984.

\bibitem[Biswas et~al.(2019)Biswas, Jacob, and Vanetti]{biswas2019estimating}
Niloy Biswas, Pierre~E Jacob, and Paul Vanetti.
\newblock Estimating convergence of markov chains with l-lag couplings.
\newblock \emph{In NeurIPS}, 2019.

\bibitem[Bonet et~al.(2021)Bonet, Courty, Septier, and
  Drumetz]{bonet2021sliced}
Cl{\'e}ment Bonet, Nicolas Courty, Fran{\c{c}}ois Septier, and Lucas Drumetz.
\newblock Sliced-wasserstein gradient flows.
\newblock \emph{arXiv preprint arXiv:2110.10972}, 2021.

\bibitem[Bonneel et~al.(2015)Bonneel, Rabin, Peyr{\'e}, and
  Pfister]{bonneel2015sliced}
Nicolas Bonneel, Julien Rabin, Gabriel Peyr{\'e}, and Hanspeter Pfister.
\newblock Sliced and radon wasserstein barycenters of measures.
\newblock \emph{Journal of Mathematical Imaging and Vision}, 51\penalty0
  (1):\penalty0 22--45, 2015.

\bibitem[Bonnotte(2013)]{bonnotte2013unidimensional}
Nicolas Bonnotte.
\newblock \emph{Unidimensional and evolution methods for optimal
  transportation}.
\newblock PhD thesis, Paris 11, 2013.

\bibitem[Bornschein and Bengio(2014)]{bornschein2014reweighted}
J{\"o}rg Bornschein and Yoshua Bengio.
\newblock Reweighted wake-sleep.
\newblock \emph{arXiv preprint arXiv:1406.2751}, 2014.

\bibitem[Cuturi(2013)]{cuturi2013sinkhorn}
Marco Cuturi.
\newblock Sinkhorn distances: Lightspeed computation of optimal transport.
\newblock In \emph{NeurIPS}, 2013.

\bibitem[Deshpande et~al.(2018)Deshpande, Zhang, and
  Schwing]{deshpande2018generative}
Ishan Deshpande, Ziyu Zhang, and Alexander~G Schwing.
\newblock Generative modeling using the sliced wasserstein distance.
\newblock In \emph{CVPR}, 2018.

\bibitem[Dieng et~al.(2017)Dieng, Tran, Ranganath, Paisley, and
  Blei]{dieng2017variational}
Adji~Bousso Dieng, Dustin Tran, Rajesh Ranganath, John Paisley, and David Blei.
\newblock Variational inference via $\chi^2 $ upper bound minimization.
\newblock \emph{In NeurIPS}, 2017.

\bibitem[Feng et~al.(2017)Feng, Wang, and Liu]{feng2017learning}
Yihao Feng, Dilin Wang, and Qiang Liu.
\newblock Learning to draw samples with amortized stein variational gradient
  descent.
\newblock In \emph{UAI}, 2017.

\bibitem[Gershman and Goodman(2014)]{gershman2014amortized}
Samuel Gershman and Noah Goodman.
\newblock Amortized inference in probabilistic reasoning.
\newblock In \emph{Proceedings of the annual meeting of the cognitive science
  society}, volume~36, 2014.

\bibitem[Gulrajani et~al.(2017)Gulrajani, Ahmed, Arjovsky, Dumoulin, and
  Courville]{gulrajani2017improved}
Ishaan Gulrajani, Faruk Ahmed, Martin Arjovsky, Vincent Dumoulin, and Aaron
  Courville.
\newblock Improved training of wasserstein gans.
\newblock In \emph{NeurIPS}, 2017.

\bibitem[Hoffman(2017)]{hoffman2017learning}
Matthew~D Hoffman.
\newblock Learning deep latent gaussian models with markov chain monte carlo.
\newblock In \emph{ICML}, 2017.

\bibitem[Jerfel et~al.(2021)Jerfel, Wang, Fannjiang, Heller, Ma, and
  Jordan]{jerfel2021variational}
Ghassen Jerfel, Serena Wang, Clara Fannjiang, Katherine~A Heller, Yian Ma, and
  Michael~I Jordan.
\newblock Variational refinement for importance sampling using the forward
  kullback-leibler divergence.
\newblock \emph{arXiv preprint arXiv:2106.15980}, 2021.

\bibitem[Jordan et~al.(1998)Jordan, Kinderlehrer, and
  Otto]{jordan1998variational}
Richard Jordan, David Kinderlehrer, and Felix Otto.
\newblock The variational formulation of the fokker--planck equation.
\newblock \emph{SIAM journal on mathematical analysis}, 29\penalty0
  (1):\penalty0 1--17, 1998.

\bibitem[Kingma and Welling(2014)]{kingma2013auto}
Diederik~P Kingma and Max Welling.
\newblock Auto-encoding variational bayes.
\newblock In \emph{ICLR}, 2014.

\bibitem[Kolouri et~al.(2018{\natexlab{a}})Kolouri, Pope, Martin, and
  Rohde]{kolouri2018sliced2}
Soheil Kolouri, Phillip~E Pope, Charles~E Martin, and Gustavo~K Rohde.
\newblock Sliced wasserstein auto-encoders.
\newblock In \emph{ICLR}, 2018{\natexlab{a}}.

\bibitem[Kolouri et~al.(2018{\natexlab{b}})Kolouri, Rohde, and
  Hoffmann]{kolouri2018sliced}
Soheil Kolouri, Gustavo~K Rohde, and Heiko Hoffmann.
\newblock Sliced wasserstein distance for learning gaussian mixture models.
\newblock In \emph{CVPR}, pages 3427--3436, 2018{\natexlab{b}}.

\bibitem[Li and Turner(2016)]{li2016renyi}
Yingzhen Li and Richard~E Turner.
\newblock R{\'e}nyi divergence variational inference.
\newblock \emph{In NeurIPS}, 29, 2016.

\bibitem[Li et~al.(2017)Li, Turner, and Liu]{li2017approximate}
Yingzhen Li, Richard~E Turner, and Qiang Liu.
\newblock Approximate inference with amortised mcmc.
\newblock \emph{arXiv preprint arXiv:1702.08343}, 2017.

\bibitem[Liu et~al.(2019)Liu, Zhuo, and Zhu]{liu2019understanding}
Chang Liu, Jingwei Zhuo, and Jun Zhu.
\newblock Understanding mcmc dynamics as flows on the wasserstein space.
\newblock In \emph{ICML}, 2019.

\bibitem[Liu and Wang(2016)]{liu2016stein}
Qiang Liu and Dilin Wang.
\newblock Stein variational gradient descent: A general purpose bayesian
  inference algorithm.
\newblock In \emph{NeurIPS}, 2016.

\bibitem[Mescheder et~al.(2017)Mescheder, Nowozin, and
  Geiger]{mescheder2017adversarial}
Lars Mescheder, Sebastian Nowozin, and Andreas Geiger.
\newblock Adversarial variational bayes: Unifying variational autoencoders and
  generative adversarial networks.
\newblock In \emph{ICML}, 2017.

\bibitem[Nadjahi et~al.(2019)Nadjahi, Durmus, Simsekli, and
  Badeau]{nadjahi2019asymptotic}
Kimia Nadjahi, Alain Durmus, Umut Simsekli, and Roland Badeau.
\newblock Asymptotic guarantees for learning generative models with the
  sliced-wasserstein distance.
\newblock \emph{In NeurIPS}, 2019.

\bibitem[Nadjahi et~al.(2020)Nadjahi, Durmus, Chizat, Kolouri, Shahrampour, and
  Simsekli]{nadjahi2020statistical}
Kimia Nadjahi, Alain Durmus, L{\'e}na{\"\i}c Chizat, Soheil Kolouri, Shahin
  Shahrampour, and Umut Simsekli.
\newblock Statistical and topological properties of sliced probability
  divergences.
\newblock \emph{In NeurIPS}, 2020.

\bibitem[Naesseth et~al.(2020)Naesseth, Lindsten, and
  Blei]{naesseth2020markovian}
Christian Naesseth, Fredrik Lindsten, and David Blei.
\newblock Markovian score climbing: Variational inference with $kl(p|| q)$.
\newblock \emph{In NeurIPS}, 2020.

\bibitem[Neal(2011)]{brooks2011handbook}
Radford Neal.
\newblock \emph{MCMC using Hamiltonian dynamics}.
\newblock CRC press, 2011.

\bibitem[Neal et~al.(2011)]{neal2011mcmc}
Radford~M Neal et~al.
\newblock Mcmc using hamiltonian dynamics.
\newblock \emph{Handbook of markov chain monte carlo}, 2\penalty0
  (11):\penalty0 2, 2011.

\bibitem[Paisley et~al.(2012)Paisley, Blei, and Jordan]{paisley2012variational}
John Paisley, David Blei, and Michael Jordan.
\newblock Variational bayesian inference with stochastic search.
\newblock In \emph{ICML}, 2012.

\bibitem[Prangle(2019)]{prangle2019distilling}
Dennis Prangle.
\newblock Distilling importance sampling.
\newblock \emph{arXiv preprint arXiv:1910.03632}, 2019.

\bibitem[Ranganath et~al.(2014)Ranganath, Gerrish, and
  Blei]{ranganath2014black}
Rajesh Ranganath, Sean Gerrish, and David Blei.
\newblock Black box variational inference.
\newblock In \emph{AIstats}, 2014.

\bibitem[Ranganath et~al.(2016)Ranganath, Tran, Altosaar, and
  Blei]{ranganath2016operator}
Rajesh Ranganath, Dustin Tran, Jaan Altosaar, and David Blei.
\newblock Operator variational inference.
\newblock \emph{In NeurIPS}, 2016.

\bibitem[Rezende and Mohamed(2015)]{rezende2015variational}
Danilo Rezende and Shakir Mohamed.
\newblock Variational inference with normalizing flows.
\newblock In \emph{ICML}, 2015.

\bibitem[Rezende et~al.(2014)Rezende, Mohamed, and
  Wierstra]{rezende2014stochastic}
Danilo~Jimenez Rezende, Shakir Mohamed, and Daan Wierstra.
\newblock Stochastic backpropagation and approximate inference in deep
  generative models.
\newblock In \emph{ICML}, 2014.

\bibitem[Ruiz and Titsias(2019)]{ruiz2019contrastive}
Francisco Ruiz and Michalis Titsias.
\newblock A contrastive divergence for combining variational inference and
  mcmc.
\newblock In \emph{ICML}, 2019.

\bibitem[Villani(2009)]{villani2009optimal}
C{\'e}dric Villani.
\newblock \emph{Optimal transport: old and new}, volume 338.
\newblock Springer, 2009.

\bibitem[Wan et~al.(2020)Wan, Li, and Hovakimyan]{wan2020f}
Neng Wan, Dapeng Li, and Naira Hovakimyan.
\newblock f-divergence variational inference.
\newblock \emph{In NeurIPS}, 2020.

\bibitem[Welling and Teh(2011)]{welling2011bayesian}
Max Welling and Yee~W Teh.
\newblock Bayesian learning via stochastic gradient langevin dynamics.
\newblock In \emph{ICML}, 2011.

\bibitem[Wolfowitz(1957)]{wolfowitz1957minimum}
Jacob Wolfowitz.
\newblock The minimum distance method.
\newblock \emph{The Annals of Mathematical Statistics}, pages 75--88, 1957.

\bibitem[Zhang et~al.(2020)Zhang, Zheng, and Zhou]{zhang2020mcmc}
Quan Zhang, Huangjie Zheng, and Mingyuan Zhou.
\newblock Mcmc-interactive variational inference.
\newblock \emph{arXiv preprint arXiv:2010.02029}, 2020.

\end{thebibliography}

\appendix

\section{Proof of Theorem 2}\label{apd:first}

\begin{proof}
First we introduce lower semi continuity of sliced Wasserstein distance (Lemma 1 of \cite{nadjahi2019asymptotic}). For any $\{\mu^n_t\}_{n\in \mathbb{N}}, \{\upsilon^n_\phi\}_{n\in \mathbb{N}} \in \mathcal{P}_p(\mathcal{X})$ weakly converge to $\mu_t, \upsilon_\phi$, the following inequality holds $\lin \sw(\mu^n_t, \upsilon^n_\phi) \geq \sw(\mu_t, \upsilon_\phi)$. Next we introduce \textbf{Lemma 4} that gives the sufficient condition that minimum exists.

\begin{lemma} $\mathcal{X}$ is a compact set and $f:\mathcal{X} \to \mathbb{R}$ is lower semi continuous, then $f$ is bounded below and it attains the infimum.
\end{lemma} \label{exist-lemma}
$B_{\epsilon} = \{\phi \in \mathcal{H}: \sw(\mu_t, \upsilon_\phi) \leq \epsilon^* + \epsilon \}$ is closed since $\phi \to \sw(\mu_t, \upsilon_\phi)$ is lower semi-continuous and By \textbf{A.4.} we have that $B_{\epsilon}$ is bounded therefore $B_{\epsilon}$ is a compact set. Hence by Lemma 4 we know $\arg\min_{\phi \in \mathcal{H}}\sw(\mu_t, \upsilon_\phi) $ is non empty. \\%i.e., $\min_{\phi \in \mathcal{H}}\sw(\mu_t, \upsilon_\phi) = \inf_{\phi \in \mathcal{H}}\sw(\mu_t, \upsilon_\phi)$.\\
\\
Next we prove that $ \lim_{n \to \infty} \inf_{\phi} \sw(\mu^n_t, \upsilon^n_\phi )  = \inf_{\phi} \sw(\mu_t, \upsilon_\phi )$, the key step is to prove the epi-convergence in sub compact set $\mathcal{K} \in \mathcal{H}$ and open set $\mathcal{O} \in \mathcal{H}$.
\begin{align}
\left\{ \begin{array}{cc} 
               &\lin \inf_{\phi \in \mathcal{K}} \sw(\mu^n_t, \upsilon^n_\phi) \geq \inf_{\phi \in \mathcal{K}} \sw(\mu_t, \upsilon_\phi) \\
                &\lsn \inf_{\phi \in \mathcal{O}} \sw(\mu^n_t, \upsilon^n_\phi) \leq \inf_{\phi \in \mathcal{O}} \sw(\mu_t, \upsilon_\phi) \\
                \end{array} \right.
\end{align}
For compact set $\mathcal{K}$, by definition such that every sequence in $\mathcal{K}$ has a convergent sub-sequence. By lower semi continuity of $\phi \to \sw(\mu^n_t, \upsilon^n_{\phi})$, %we know that any continuous function on a compact set is bounded. 
we have 
$
    \inf_{\phi \in \mathcal{K}} \sw(\mu^n_t, \upsilon^n_\phi) = \sw(\mu^n_t, \upsilon^n_{\phi_n})
$ with some $\phi_n \in \mathcal{K}$. Hence
\begin{equation}
\begin{split}
   &\lin \inf_{\phi \in \mathcal{K}} \sw(\mu^n_t, \upsilon^n_\phi) = \lin \sw(\mu^n_t, \upsilon^n_{\phi_n}) \\
   & = \lim_{k \to \infty }\sw(\mu^{n_k}_t, \upsilon^{n_k}_{\phi_{n_k}}) \text{, exists sub-sequence converges to }\lim\inf \\
   & = \lim_{l \to \infty }\sw(\mu^{n_{k_l}}_t, \upsilon^{n_{k_l}}_{\phi_{n_{k_l}}})  \\
   & = \lil \sw(\mu^{n_{k_l}}_t, \upsilon^{n_{k_l}}_{\phi_{n_{k_l}}}) \\
   & \geq \lil \left[\sw(\mu_t, \upsilon^{n_{k_l}}_{\phi_{n_{k_l}}})  -  \sw(\mu^{n_{k_l}}_t, \mu_t)  \right], \text{by triangular inequality} \\
   & \geq \lil \sw(\mu_t, \upsilon^{n_{k_l}}_{\phi_{n_{k_l}}}) -   \lsl \sw(\mu^{n_{k_l}}_t, \mu_t)\\ 
   & \geq \sw(\mu_t, \upsilon_{\bar{\phi}}) \text{, by Assumption \textbf{A.3.} and exists sub-sequence converges in $\mathcal{K}$ to $\bar{\phi}$}\\
   & \geq \inf_{\phi \in \mathcal{K}} \sw(\mu_t, \upsilon_\phi)
  \end{split}
\end{equation}
For open set $\mathcal{O} \in \mathcal{H}$, there exists $\{\phi_n\} \in \mathcal{O}$ such that $\lim_{n \to \infty} \sw(\mu_t, \upsilon_{\phi_n}) = \inf_{\phi \in \mathcal{O}} \sw(\mu_t, \upsilon_{\phi})$. And $\forall
 n \in \mathbb{N}, \sw(\mu^n_t, \upsilon_{\phi_n}) \geq \inf_{\phi \in \mathcal{O}} \sw(\mu^n_t, \upsilon_{\phi}) $. Hence
\begin{equation}
\begin{split}
   &\lsn \inf_{\phi \in \mathcal{O}} \sw(\mu^n_t, \upsilon^n_\phi) \leq \lsn \sw(\mu^n_t, \upsilon^n_{\phi_n}) \\
   & \leq \lsn \left[ \sw(\mu^n_t, \mu_t) + \sw(\mu_t, \upsilon_{\phi_n}) + \sw(\upsilon_{\phi_n}, \upsilon^n_{\phi_n})\right], \text{by triangular inequality} \\
   & \leq \lsn \sw(\mu^n_t, \mu_t) + \lsn \sw(\mu_t, \upsilon_{\phi_n}) + \lsn\sw(\upsilon_{\phi_n}, \upsilon^n_{\phi_n}) ,\text{by boundedness} \\
   &=\lsn\sw(\mu_t, \upsilon_{\phi_n}), \text{ by Assumption \textbf{A.3.}} \\
   &= \inf_{\phi \in \mathcal{O}} \sw(\mu_t, \upsilon_\phi)
  \end{split}
\end{equation}
Hence we have derived the epi-convergence. It is trivial to prove the convergence under $\mathcal{H}$ $\lim_{n \to \infty} \inf_{\phi \in \mathcal{H}} \sw(\mu^n_t, \upsilon^n_\phi )  = \inf_{\phi \in \mathcal{H}} \sw(\mu_t, \upsilon_\phi)$. We refer to \citep{bernton2019parameter, nadjahi2019asymptotic} for similar derivations.
\end{proof}

\section{Experiment Settings}\label{apd:second}
\textbf{4.1.} Target distributions:
\scriptsize{
 \begin{align*}
 \text{2D Gaussian: } 
 N
\begin{bmatrix}
\begin{pmatrix}
1\\
2
\end{pmatrix},
\begin{pmatrix}
1.00 & 0.80 \\
0.80 & 0.92 
\end{pmatrix}
\end{bmatrix}
\text{,   Mixture: } 
0.5N
\begin{bmatrix}
\begin{pmatrix}
-1\\
1
\end{pmatrix},
\begin{pmatrix}
0.80 & 0.00 \\
0.00 & 0.80 
\end{pmatrix}
\end{bmatrix}
+0.5N
\begin{bmatrix}
\begin{pmatrix}
3\\
-3
\end{pmatrix},
\begin{pmatrix}
0.80 & 0.00 \\
0.00 & 0.80 
\end{pmatrix}
\end{bmatrix}
\end{align*}}\\
\normalsize All variational distributions are initialized at 
\scriptsize{
 \begin{align*}
 N
\begin{bmatrix}
\begin{pmatrix}
-0.5\\
-0.5
\end{pmatrix},
\begin{pmatrix}
0.50 & 0.00 \\
0.00 & 0.50 
\end{pmatrix}
\end{bmatrix}
\end{align*}}
\normalsize The number of Markov chains is 300 and starts from the initial variational distribution with lag $L=20$. The same number of samples are used in VI for reparameterizing gradient estimation. \\
\\
\textbf{4.2.} Target distributions:
\scriptsize{
 \begin{align*}
0.5N
\begin{bmatrix}
\begin{pmatrix}
1\\
1
\end{pmatrix},
\begin{pmatrix}
1.00& 0.00 \\
0.00 & 1.00 
\end{pmatrix}
\end{bmatrix}
+0.5N
\begin{bmatrix}
\begin{pmatrix}
-1\\
-2
\end{pmatrix},
\begin{pmatrix}
0.04 & 0.00 \\
0.00 & 0.04 
\end{pmatrix}
\end{bmatrix}
\end{align*}}\\
\normalsize Neural networks have 3 hidden layers (each with 128 units) with input dimension 5. The number of Markov chains is 300 and starts at samples output by the initial Neural networks. Lag $L=20$. MCMC is random walk Metropolis-Hastings with step size 2.0. SVGD employs 300 particles and uses RBF kernel with empirical median trick bandwidth \citep{liu2016stein}.\\
\\
\textbf{4.4.} We pre-train a varitional auto-encoder (VAE) on MNIST using 1,200 randomly selected training samples.  The encoder of VAE has one hidden layer with 100 units for both mean and log-variance. The decoder is also one hidden layer neural network with 100 units and with sigmoid activation for the output layer. The latent variable has 10 dimensions.\\
\\
We fix decoder pre-trained with VAE and set prior distribution to standard Gaussian such that posterior in Eq (18) is defined. 100 randomly selected samples from test set are used for evaluation of posterior approximations. For 'Neural net SWVI', we use a two hidden neural net with units $[128, 64]$ and the input dimension is 5. We parallel 10 HMC chains and HMC takes 3 steps leapfrog with step size 0.1. All HMC chains start at samples generated from the learned mean-field Gaussian via VI. we run HMC 10 times for 'Mean-field VI+HMC', 'Mean-field SWVI' and 100 times for 'Neural net SWVI' (lag $L=0$ in this experiment).\\\\
\textbf{Remark} The number of slices sampled for estimating sliced Wasserstein distance in all experiments is 10 excluding \textbf{4.4} where we use 100 slices for 'Neural net SWVI' .

\end{document}